\documentclass[runningheads,a4paper]{llncs}
\pdfoutput=1
\usepackage{amssymb}
\usepackage{graphicx}
\usepackage{amsmath}
\usepackage{url}
\urldef{\mailsc}\path|gabriel.murray@ufv.ca|    
\newcommand{\keywords}[1]{\par\addvspace\baselineskip
\noindent\keywordname\enspace\ignorespaces#1}

\usepackage[hidelinks]{hyperref}
\hypersetup{breaklinks=true}

\usepackage{cite}

\begin{document}

\mainmatter  

\title{Stoic Ethics for Artificial Agents}

\titlerunning{Stoic Ethics for Artificial Agents}

%
%
\author{Gabriel Murray}

\authorrunning{Gabriel Murray}

\institute{University of the Fraser Valley, Abbotsford, BC, Canada\\
\mailsc\\
\url{http://www.ufv.ca/cis/gabriel-murray/}}

\maketitle

\begin{abstract}

We present a position paper advocating the notion that Stoic philosophy and ethics can inform the development of ethical A.I. systems. This is in sharp contrast to most work on building ethical A.I., which has focused on Utilitarian or Deontological ethical theories. We relate ethical A.I. to several core Stoic notions, including the dichotomy of control, the four cardinal virtues, the ideal Sage, Stoic practices, and Stoic perspectives on emotion or affect. More generally, we put forward an ethical view of A.I. that focuses more on internal states of the artificial agent rather than on external actions of the agent. We provide examples relating to near-term A.I. systems as well as hypothetical superintelligent agents. 

\keywords{ethical A.I., virtue ethics, Stoicism, superintelligence}
\end{abstract}

\section{Introduction}
\label{sec:intro}

Stoicism is a philosophy that was prominent during the Hellenistic period and into the era of the Roman Empire \cite{Pigliucci}. Its ethical view is a form of Virtue Ethics, and both Stoicism and Virtue Ethics have seen a resurgence in study and popularity in recent decades. Virtue Ethics is now one of the three major ethical perspectives, alongside Utilitarianism and Deontological Ethics \cite{Hursthouse2016}. Whereas Utilitarianism examines the utility of actions and their consequences, and Deontological Ethics studies duties and obligations, Virtue Ethics focuses on virtuous or moral character, happiness, and the good life \cite{Sandel2010}. Stoicism more specifically adds ideas and practices relating to emotion, control, and rational deliberation, all of which we consider in depth.

The importance of designing ethical A.I. systems has become widely recognized in recent years. Most of this work is influenced by Utilitarian and Deontological ethics; for example, proposing or designing intelligent systems that act in such a way as to maximize some aspect of human well-being, or to obey particular rules. There has been very little work on how Virtue Ethics can inform the development of ethical A.I. agents, and perhaps no such work focusing on Stoicism. This is not surprising, as it is not immediately clear what it would even mean for an artificial agent to be Stoic. In this position paper, we aim to describe concretely what that could mean, and we advocate that Stoicism and Virtue Ethics should be considered in the discussion around ethical A.I. 

We propose that Stoic ethical analysis of A.I. should include the following:
\begin{itemize}
\item Stoic ethical analysis of A.I. should be based on analysis of an agent's internal states. 
\item Stoic ethical judgment of an A.I. agent should not be based on factors outside of the agent's control, and specifically should not be consequentialist.
\item Stoic ethical analysis of A.I. should include whether or not the agent has characteristics corresponding to Stoic virtues. 
\end{itemize}

Similarly, A.I. systems should be \textit{designed} with these considerations in mind, and incorporate aspects of Stoic practices. 

In Section \ref{sec:related}, we discuss current directions in ethical A.I., including the limited previous work on Virtue Ethics for A.I. systems. In Section \ref{sec:stoicai}, we discuss some of the core concepts of Stoicism and how they can relate to the development of A.I. systems. These concepts include Stoic control, the four cardinal virtues, the ideal Sage, emotion, and Stoic practices. In Section \ref{sec:critic}, we consider criticisms of Stoicism and Virtue Ethics, as well as criticism of the notion that they could have any bearing on designing ethical intelligent systems. In that section, we also give some limited proposals on how Stoicism can be combined with Utilitarian and Deontological considerations. Finally, we conclude and summarize our position in Section \ref{sec:conclusion}.

\section{Related Work}
\label{sec:related}

Recent discussion on ethical A.I. can be divided into that which focuses primarily on ethical implication of current and near-term A.I. systems, and that which primarily focuses on hypothetical superintelligence, i.e. systems with intelligence and capabilities that are beyond human-level. As an example of the former, Amodei et al. \cite{Amodei2016} describe concrete problems relating to A.I. safety, and Arkin \cite{Arkin2009} discusses ethics for autonomous systems used in warfare. Wallach and Allen \cite{Wallach2008} discuss morality for agents of varying levels of sophistication. Bostrom \cite{Bostrom2014}, Dewey \cite{Dewey2011}, Christiano \cite{Christiano2014}, and Yudkowsky \cite{Yudkowsky2004} focus on A.I. ethics with a particular emphasis on superintelligent systems. Much of this work, whether looking at near-term or long-term A.I., is based on reinforcement learning, where an artificial agent learns how to act in an environment by collecting rewards (which could be negative, i.e. punishments). There are well-recognized potential pitfalls with this approach, such as using the wrong reward function, or having a superintelligent agent take control of its own reward signal (called reward hacking, or wireheading). We will discuss some of these details and issues in Section \ref{sec:stoicai}. In general, these approaches are action-centric and based on taking actions that maximize some reward or utility. 

There has been very little work on applying Virtue Ethics to artificial intelligence. Coleman \cite{Coleman2001} describes characteristics that can help an A.I. achieve its goals, and analyzes these characteristics as virtues in a Virtue Ethic framework. This is similar to our effort, except that we focus on the four cardinal Stoic virtues, and discuss many additional Stoic ideas and practices as well. Hicks \cite{Hicks2014} discusses drawbacks to designing ethical A.I. using principle-based approaches such as Utilitarian or Deontological ethics, and argues in favour of a role for Virtue Ethics. As one example, Hicks argues that in scenarios where all options are bad but the A.I. agent must still make a decision, the agent should have some sense of regret about the decision. 

As the field of artificial intelligence has developed, the topic of ethical A.I. has been relatively neglected. The most famous example of ethical A.I. has come from fiction rather than A.I. research, in the form of Asimov's Laws of Robotics \cite{wiki:asimov}. Recent years have seen a marked increase in attention paid to the topic of ethical A.I., with annual workshops as well as organizations and partnerships dedicated to the matter \footnote{\url{https://www.partnershiponai.org/} \\ \url{https://intelligence.org/} \\ \url{http://futureoflife.org/} \\ \url{http://humancompatible.ai} \\ \url{http://lcfi.ac.uk/}}.

\section{Stoicism and A.I.}
\label{sec:stoicai}

In this section we discuss how Stoic philosophy and ethics can be relevant to the development of ethical A.I. systems. 

\subsection{Internal States Matter}

As mentioned in Section \ref{sec:related}, most existing work on ethical A.I. is action-centric. For example, we can design an intelligent agent that tries to maximize a reward function relating to some aspect of human well-being. There is a clear parallel between such systems and the Utilitarian ethical perspective, which says that an action is good if its positive consequences outweigh its negative consequences. Alternatively, an action-centric system could be informed by Deontological ethics, so that it takes actions that do not violate its obligations, and that do not use people or other agents as a means to an end (i.e. it acts in accordance with the Categorical Imperative). In either case, actions and consequences external to the agent are of paramount importance.

Stoicism and Virtue Ethics are more concerned with internal states of the agent. In that light, we need to first make a case that the internal states of an A.I. agent matter. We will give a number of examples demonstrating that the internal states of an A.I. do indeed matter, starting with current and near-term A.I. and then moving to increasingly sophisticated A.I. and hypothetical superintelligent systems. 

A first example of the importance of internal states of an A.I. is the growing recognition that an A.I. system should be as \textit{transparent} and \textit{explainable} as possible\footnote{\url{http://www.fatml.org/}}. If an A.I. system makes a prediction or a decision, we should be able to perform an audit that determines why it behaved as it did, or issue a query about what it has learned. A second, related example is that A.I. systems should be \textit{corrigible}, or amenable to correction in order to improve behaviour and reduce mistakes \cite{Soares2015}. 

Another example concerns intelligent systems with affective capabilities. Affective computing involves the ability of a system to recognize, represent, and/or simulate affective states or emotions \cite{Calvo2014}. Since the role of emotion is critical in Stoic ethics, we will have much more to say about it below. At this point, it suffices to say that the internal states of an intelligent system with affective capabilities are important because the system may be engaging in persuasion or emotional manipulation that may be subtle from an external perspective. 

If and when an A.I. system begins to match or exceed general human intelligence, we may need to engage in thorough monitoring of the system's internal states to carefully assess the agent's progress and capabilities. If a system is superintelligent but is a \textit{boxed} A.I., meaning that it has severe restrictions on its external capabilities, it is possible that the agent will be strategically cooperative until it is unboxed, at which point it will have a strategic advantage and no longer need to cooperate with humans. It may also use emotional manipulation to persuade human operators to unbox it \cite{Bostrom2014}. 

The prospect of such superintelligent systems leads to what Bostrom calls \textit{mind crimes}, i.e. immoral internal states. The example just given, of an A.I. being deceptively cooperative, could constitute an immoral internal state. Bostrom gives a more dramatic example, where a superintelligent agent is able to create conscious human simulations, and is moreover able to create and destroy such conscious simulations by the billions. If the human simulations are truly conscious, this would constitute massive genocide \cite{Bostrom2014}. 

These examples show that internal states are important even with current and near-term A.I. systems, and could become more so if superintelligent systems are ever developed.

\subsection{Control}

A central notion of Stoicism is the \textit{dichotomy of control}, or recognizing that some things are in our control and some things are not. A person should not be overly concerned with things outside of their control, and more importantly, things that are outside of a person's control are neither good nor bad, morally speaking. Only virtue is good, and being virtuous is always within a person's control. And even when dealing with events that are outside of our control, we still control our responses to those events.

There are several applications of this idea to artificial intelligence. Clearly, it is important for an A.I. agent to know what is and isn't under its control, i.e. what are its capabilities and what is the environment like. And in an uncertain environment, the agent still controls its responses. Modern Stoic thinkers have used the idea of a \textit{trichotomy of control} \cite{Irvine2008}, introducing a third category of things we partly control, and that category can be demonstrated with two examples from within artificial intelligence. To use the example of an Markov Decision Process (MDP) with an optimal policy, the agent is best off following the policy even though the dynamics are uncertain (it controls its choice of actions, but not the dynamics and environment). To use a second example of a multi-agent scenario with two players and a zero-sum game, the first agent can adopt a \textit{minimax} strategy to minimize its maximum loss (it controls its choice, but not the choice of the other agent). 

More importantly, we should not say that an A.I. agent is exhibiting morally wrong behaviour if that judgment is based on factors that are outside of its control. This distinguishes Virtue Ethics from Act Utilitarianism, with the latter being susceptible to the problem of \textit{moral luck}. Moral luck is a problem for consequentialist theories of ethics, because unforeseen circumstances can lead to disastrous consequences, which would cause us to determine that the action was wrong.\footnote{Rule Utilitarianism addresses the problem of moral luck, and Eric O. Scott points out (personal communication) that a consequentialist analysis of an agent can also address the problem by considering a large number of trials.}
In contrast, Virtue Ethics says that the agent is behaving correctly if it is behaving in accordance with the cardinal virtues, regardless of external consequences.

\subsection{Affective Computing}

The Stoic notion of control is closely related to their perspective on emotions (or ``passions''). Because the word \textit{stoic} has entered the general vocabulary as meaning \textit{lacking emotion}, Stoics have been wrongly thought of as being unemotional or disdaining emotion. In fact, their perspective on emotion is that we control our emotional responses to events even if we do not control the events themselves, and that it is pointless to fret about things outside of our control. A person obsessing about things outside their control is needlessly suffering. 

Recall that affective capabilities for an intelligent system involve the ability to recognize, represent, and/or simulate emotions or affective states \cite{Calvo2014}. Let us call an agent with such capabilities an Affective Agent. Just as a Stoic person can monitor and control their emotional responses to events, so can an Affective Agent monitor and control how it represents and simulates emotional states. This has implications not just for the agent, but also for humans interacting with the agent. At one extreme, a person could be emotionally manipulated by an Affective Agent, while at the other extreme the Affective Agent could be a calming influence on a person. 

We briefly note that even a system without explicit affective capabilities can be a calming influence. Christian and Griffiths \cite{Christian2016} introduce the notion of \textit{Computational Stoicism} to refer to the peace of mind that comes with using optimal algorithms in our everyday lives. If we are employing an optimal algorithm to solve some problem, then we can do no better using the tools that are under our control, regardless of what consequences may result. 

We need to be careful about describing an A.I. as Stoic and what that means with regard to emotion, as it is very likely that many people fear the prospect of a generally intelligent A.I. that is devoid of emotion and operates in a cold, calculating manner. Just as it is a mistake to describe a Stoic person as emotionless, it would be wrong to describe a Stoic A.I. as being emotionless. A Stoic A.I. would be an A.I. that is in accord with Stoic virtues, and has carefully calibrated affective capabilities. Stoic A.I. is virtuous A.I.

\subsection{Stoic Virtues}

We now turn to the four cardinal Stoic virtues of wisdom, justice, courage, and temperance, and how they relate to characteristics of A.I. agents.

\paragraph{\textbf{Wisdom}:} It may seem obvious that an intelligent agent should have wisdom, but it is not necessarily the case that even an advanced A.I. would be wise, since intelligence and wisdom are not normally considered by humans to be synonymous. That an advanced A.I. could lack wisdom becomes evident when we examine the virtues that are categorized under wisdom: good sense, good calculation, quick-wittedness, discretion, and resourcefulness \cite{Stephens,Pigliucci}. Computers are already beyond human-level in terms of excelling at calculation and being very fast. However, discretion and good sense are qualities that we do not necessarily associate with existing A.I. systems. Discretion for an intelligent agent could mean respecting privacy, and preserving anonymity in data. Good sense could incorporate commonsense reasoning, while resourcefulness could mean being innovative in finding new solutions to a problem.

The Stoics had three main areas of study: ethics, physics, and logic. They believed the three areas to be interdependent, because you cannot determine right or wrong (ethics) without understanding the world (physics) and thinking rationally (logic). A wise A.I. would need to be able to reason about real-world knowledge, including commonsense reasoning, while also being innovative and flexible -- and A.I. systems based purely on logic are not always flexible. The Stoics also advocated the idea of being guided by the wisdom of an \textit{Ideal Sage}, which we will discuss further below.

\paragraph{\textbf{Justice}:} The virtues categorized under justice include piety, honesty, equity, and fair dealing \cite{Stephens,Pigliucci}. An example of a just A.I. agent is one that does not mislead, even if the consequences of misleading could benefit the agent. The virtue of piety could be realized by an A.I. that is in a principal-agent relationship where it is designated to act on behalf of some principal, in which case it should be devoted to the principal and not to its own ends. Equity and fair dealing mean that an A.I. agent should not be biased in terms of race, gender, sexual orientation, or socioeconomic status, and should not use individuals as a means to its own ends.

\paragraph{\textbf{Courage}:} Courage is often thought of as bravery, and we can easily imagine, for example, a military A.I. that is willing to be destroyed in order to save human lives. But when Stoic philosophers (and other Virtue Ethicists) talk about courage, they are speaking primarily about \textit{moral courage}, which means taking some action for moral reasons, despite the consequences. So again we see a rejection of consequentialism, this time in the form of Stoic moral courage.

The virtues categorized under courage include endurance, confidence, high-mindedness, cheerfulness, and industriousness \cite{Stephens,Pigliucci}. Confidence can mean that the A.I. agent is a good Bayesian that is able to update its prior beliefs as it gathers evidence. It does \textit{not} mean that the A.I. agent has confidence at all times, even when unwarranted. In fact, being explicit about its uncertainty and recognizing that it could be wrong can be seen as a virtue of humility that is closely coupled with the virtue of confidence.

\paragraph{\textbf{Temperance}:} The virtues included under temperance include good discipline, seemliness, modesty, and self-control \cite{Stephens,Pigliucci} . Overall, temperance involves restraint on the part of the A.I. agent. Corrigibility, or the ability to be corrected and improved, could be seen as a type of modesty. Seemliness involves a sense of decorum and appropriateness. The lack of the virtue of seemliness has been demonstrated by a simple A.I. known as Tay, a Twitter chatbot launched by Microsoft that began tweeting racist and graphic content \cite{Bright2016,Sinders2016}. Self-control can be based on feedback and learning. Bostrom describes the idea of \textit{domesticity}, where an A.I. agent has a limited scope of ambitions and activities. Bostrom also discusses the idea of instructing an A.I. agent to accomplish its goals while minimizing the impact that it has on the world \cite{Bostrom2014}. Amodei et al. \cite{Amodei2016} similarly propose either defining or learning an \textit{impact regularizer} for an agent. Temperance is an intriguing virtue for an A.I., as it seems to be at odds with the maximizing nature of many A.I. systems. We suggest that A.I. agents should rarely be engaged in pure maximization of some reward function, but instead be doing maximization subject to various constraints on their impact, or else have those constraints included in the reward function. We sketch out one such algorithm in Section \ref{sec:critic}. Alternatively, satisficing rather than maximizing models can be explored \cite{Simon1956}.

\subsection{Moral Progress and the Ideal Sage}

Stoics believe in moral progress toward some ideal virtuous state, while also recognizing that perfect virtue is not attainable. This emphasizes the need for a Stoic A.I. to be able to learn from experience, including from its own actions and mistakes, rather than being inflexible and having its behaviour predefined by strict rules. We previously mentioned the idea of an A.I. having some sense of regret when it has made a choice from a set of options that are all bad. At first, the idea of an A.I. agent with regret may seem to conflict with Stoic ideas about control, since the past is outside of our control and it is pointless to obsess about things that have already happened. While that is true, a Stoic A.I. can retain a memory of its choices and mistakes, including a sense of responsibility for things that have transpired, with an aim toward improving its behaviour in similar situations. This would be an indication of moral progress. Improving its behaviour in similar situations may entail creativity and coming up with options that are more acceptable and virtuous than the options previously known to the agent.\footnote{It should be noted, however, that creativity is a double-edged sword for an A.I. agent, since it may come up with very unexpected and disastrous solutions to a problem, such as Bostrom's extreme example of a superintelligent agent getting rid of cancer in humans by killing all humans \cite{Bostrom2014}. That is a clear example of the difference between creativity and moral progress.}

To assist in making moral progress, Stoics use the notion of an Ideal Sage, and consider what the perfectly wise and virtuous sage would do in particular scenarios. While it is usually assumed that the Ideal Sage is hypothetical and that there is no perfectly virtuous person, ancient Stoics used Socrates and Cato the Younger as two figures to consider when looking for guidance \cite{Robertson2013}. There are many ways that the concept of an Ideal Sage could be used by an A.I. agent, and we will consider them in order of increasing complexity. A simple version of an Ideal Sage is an optimal algorithm: the A.I. agent should use an optimal algorithm wherever it is possible and efficient, and when it is infeasible to use the optimal algorithm, the A.I. agent can try to approximate the optimal algorithm, or consider how the optimal algorithm would operate on a simplified version of the problem at hand. The A.I. agent may be able to calculate a competitive ratio for the algorithm it uses, describing how close it is to the performance of the optimal algorithm. A second very simple example of an Ideal Sage is in the form of gold-standard feedback provided by a human, e.g. the principal in a principal-agent relationship.

That second notion leads us towards Christiano's proposal for ``approval-directed agents,'' instead of goal-directed agents \cite{Christiano2014,Christiano2015}. An approval-directed agent takes actions that it thinks its overseer would highly rate, and can avoid issues with goal-directed agents such as misspecified goals. Christiano also emphasizes that the agent's \textit{internal} decisions can also be approval-directed. The following quote from Christiano \cite{Christiano2014} nicely parallels the idea that the Ideal Sage (in this case, a perfect overseer) represents an unattainable state but that we can start simply and scale up:

\begin{quote}
``Asking an overseer to evaluate outcomes directly requires defining an extremely intelligent overseer, one who is equipped (at least in principle) to evaluate the entire future of the universe. This is probably impractical overkill for the kinds of agents we will be building in the near future, who don’t have to think about the entire future of the universe.
Approval-directed behaviour provides a more realistic alternative: start with simple approval-directed agents and simple overseers, and scale up the overseer and the agent in parallel.'' 
\end{quote}

For example, the approval-directed agent can begin learning from examples that have been labelled by the overseer as acceptable or not acceptable behaviour. In turn, the agent can assist and improve the overseer's capabilities. Christiano's proposal represents an implementable path forward for developing virtuous agents. 

We conclude our discussion of the Ideal Sage with proposals by Bostrom and Yudkowsky for how a superintelligence could learn values. Bostrom \cite{Bostrom2014} has proposed several variants of what he calls the \textit{Hail Mary} problem, wherein the A.I. agent considers hypothetical alien superintelligences and tries to determine what those alien intelligences would likely value. Yudkowsky \cite{Yudkowsky2004} (further discussed by Bostrom \cite{Bostrom2014}) proposes that an A.I. agent can learn values by considering what he calls the coherent extrapolated volition (CEV) of humanity. The CEV essentially represents what humans would do if we were more intelligent and capable, and the A.I. agent determines its values by learning and analyzing the CEV. Both proposals have provoked much discussion but have a shared weakness, in that it is not at all clear how to implement them in the near future. 

\subsection{Stoic Practices}

In addition to the Stoic virtues and Ideal Sage, we can discuss other Stoic practices that may have interesting parallels for an A.I. agent. The first practice is the \textit{premeditatio malorum}, roughly translated as an anticipation of bad things to come \cite{Robertson2013}. For Stoic practitioners, this is often a morning meditation in which they anticipate potential misfortunes they may encounter, which could range from dealing with unkind people to losing a loved one. The idea is not to obsess over things that have not yet happened (which would be un-Stoic), but rather to rob these events of their ability to shock or control us. For an artificial agent, such planning could be used to consider worst-case scenarios, minimize maximum potential losses, and identify whether other agents are competitive or cooperative. 

Another Stoic practice is to append the phrase ``fate permitting'' to all plans and aspirations, which represents a recognition that the person controls their own actions but not the consequences. For an A.I. agent, the parallel might be that the agent should always mind the uncertainty in its plans and its environments. The agent's internal states corresponding to plans and goals may not lead to the desired external states. 

A final Stoic practice is the evening meditation, in which a person reviews the events of the day and engages in moral self-examination. For example, they may ask themselves whether they behaved virtuously and acted in accordance with the core ideas of Stoicism, or whether they could have done things differently. Again, the idea is not to obsess over things that have already happened and are therefore outside of our control, but rather to learn from experience and engage in moral progress. The parallels for an A.I. agent are clear, as any ethical agent will need to learn and improve. If the agent had been operating in a situation where it had to make choices with very limited information and time, it could later look for alternative, innovative choices that would have been more in accordance with the cardinal virtues or the preferences of an overseer. If the agent had been forced to employ a greedy algorithm while processing data in an online manner, it could later examine how performance might have improved in an optimal offline environment, and calculate the competitive ratio of the greedy algorithm it had been using. It could revisit the decisions that it had made and perform a self-audit examining the reasons behind the decision, and whether those reasons will be transparent and understandable to a human interrogator.

\section{Criticisms and Responses}
\label{sec:critic}

We can consider potential criticisms against Stoicism and the idea that it has any bearing on developing ethical A.I. A general criticism of Stoicism and other forms of Virtue Ethics is that they do not provide a strong normative ethics for real-world situations in which a person needs to make a decision given their available information \cite{Hursthouse2016}. Where Utilitarianism and Deontological ethics provide principles that are precise and (ostensibly) actionable, Stoicism gives a much vaguer principle: do what a virtuous person would do. A response to this accusation is to accept it and to point out that both Utilitarianism and Deontological Ethics are more unclear than they seem at first glance. For example, two Utilitarians can be in the same situation with the same available information and come to very different conclusions about the ethical course of action. Utilitarianism and Deontological Ethics over-promise in that they make ethical decision-making seem like a very simple application of a single rule (respectively, the principle of maximum utility and the Categorical Imperative), which can lead to overconfidence, or to a perverse application of the rule. For example, an A.I. agent that is trying to maximize a reward relating to human well-being could wind up directly stimulating the pleasure centres of the human brain in order to bring about a sense of well-being \cite{Bostrom2014}. 

An A.I. that is trying to maximize some reward function can engage in \textit{reward hacking}, also known as \textit{wireheading} \cite{Bostrom2014,Amodei2016}. This is a situation where a very advanced A.I. is able to gain control of the reward signal, so that it can get a very large reward without having to take any external actions at all. Something similar could happen with a Stoic A.I. if an actor-critic architecture is used, e.g. the critic module could be disabled. Similarly, an approval-directed A.I. could become so intelligent that it is certain it knows the overseer better than the overseer does, and thus decide to be its own overseer, granting itself approval for any action. For this reason, we encourage research into overseer hacking, just as current research examines reward hacking. This relates to the idea of \textit{adjustable autonomy} in multi-agent systems, and the opportunities and challenges associated with an agent being able to dynamically change its autonomy so that it defers to a person more often or less often \cite{Wooldridge2009}. 

Against the accusation that Stoicism is too vague for agents who need to make specific decisions and actions, we offer two responses:

\begin{itemize}
\item Approval-directed architectures offer a clear, implementable path forward, with the overseer acting as an Ideal Sage that seeds the A.I. with examples of approved and disapproved scenarios and actions. 
\item A syncretic ethics can be derived that combines elements of Stoicism with Utilitarian and Deontological ideas.
\end{itemize}

In the following subsection, we briefly describe a syncretic ethics approach for A.I. that includes Stoicism and Virtue Ethics.

\subsection{Paramedic Ethics for Artificial Agents}

Collins and Miller \cite{Collins1992} provide a ``paramedic ethics'' for Computer Science professionals who may need to take action in a variety of scenarios where time and information are limited, and where the decision-makers may not have in-depth knowledge of ethical theories. They propose a syncretic algorithm that combines elements of Utilitarianism, Deontological Ethics, and Social Contract Theory. It is possible to come up with a similar algorithm that also incorporates Virtue Ethics. For example, a very simplified and general algorithm could be structured as follows:

\begin{itemize}
\item Gather data. 
\item Determine the available actions. 
\item For each action $a$:
\begin{enumerate}
\item Does $a$ satisfy the agent's obligations (to a human principal, other agents, the law, etc.)? (the Deontological step)
\item Does $a$ accord with the cardinal virtues (or would it be approved by the overseer)? (the Stoic step)
\item What is the expected utility of $a$? (the Utilitarian step)
\end{enumerate}
\item Decide on an action by maximizing step 3 while satisfying the constraints in steps 1 and 2.
\end{itemize}

\section{Conclusion}
\label{sec:conclusion}

In this position paper, we have attempted to show how Stoic ethics could be applied to the development of ethical A.I. systems. We argued that internal states matter for ethical A.I. agents, and that internal states can be analyzed by describing the four cardinal Stoic virtues in terms of characteristics of an intelligent system. We also briefly described other Stoic practices and how they could be realized by an A.I. agent. We gave a brief sketch of how to start developing Stoic A.I. systems by creating approval-directed agents with Stoic overseers, and/or by employing a syncretic paramedic ethics algorithm with a step featuring Stoic constraints. While it can be beneficial to analyze the ethics of an A.I. agent from several different perspectives, including consequentialist perspectives, we have argued for the importance of also conducting a Stoic ethical analysis of A.I. agents, where the agent's internal states are analyzed, and moral judgments are not based on consequences outside of the agent's control.

\paragraph{Acknowledgements}  Thanks to Eric O. Scott for helpful feedback and discussion. 

\bibliographystyle{splncs}
\bibliography{stoic.bib}

\end{document}